\documentclass{esannV2}
\usepackage[dvips]{graphicx}
\usepackage[latin1]{inputenc}
\usepackage{amssymb,amsmath,array}
\usepackage{booktabs}
\usepackage{color,soul}
\usepackage{subfig}
\usepackage{graphics}

\usepackage{graphicx}
\usepackage[export]{adjustbox}

\begin{document}
\title{CNN Encoder to Reduce the Dimensionality of Data Image for Motion Planning}

\author{Janderson Ferreira$^1$, Agostinho A. F. J\'unior$^1$, Yves M. Galv\~ao$^1$ \\ Bruno J. T. Fernandes$^1$ and Pablo Barros$^{1,2}$
%
\thanks{This study was financed in part by the Coordena\c{c}\~ao de Aperfei\c{c}oamento de Pessoal de N\'ivel Superior - Brasil (CAPES) - Finance Code 001, and the Brazilian agencies FACEPE and CNPq.}
%
\vspace{.3cm}\\
%
%
1- Universidade de Pernambuco - Escola Polit\'ecnica de Pernambuco \\
Rua Benfica 455, Recife/PE - Brazil
%
\vspace{.1cm}\\
2- Cognitive Architecture for Collaborative Technologies (CONTACT) Unit \\ Istituto Italiano di Tecnologia, Genova, Italy
}


\maketitle

\begin{abstract}
Many real-world applications need path planning algorithms to solve tasks in different areas, such as social applications, autonomous cars, and tracking activities. And most importantly motion planning. Although the use of path planning is sufficient in most motion planning scenarios, they represent potential bottlenecks in large environments with dynamic changes. To tackle this problem, the number of possible routes could be reduced to make it easier for path planning algorithms to find the shortest path with less efforts. An traditional algorithm for path planning is the A*, it uses an heuristic to work faster than other solutions. In this work, we propose a CNN encoder capable of eliminating useless routes for motion planning problems, then we combine the proposed neural network output with A*. To measure the efficiency of our solution, we propose a database with different scenarios of motion planning problems. The evaluated metric is the number of the iterations to find the shortest path. The A* was compared with the CNN Encoder (proposal) with A*. In all evaluated scenarios, our solution reduced the number of iterations by more than 60\%. 
\end{abstract}

\section{Introduction}
\label{secacao_1}
Present in many daily applications, path planning algorithm helps to solve many navigation problems. Dijkstra proposed one of the first path planning algorithms in 1956 \cite{misa2010interview}. Although this algorithm was capable of finding the shortest path between two points, its use was limited to nodes with positive weights, and the computational cost is quadratic. To decrease the time complexity and allow negative weight between the nodes, others algorithms based on Dijkstra's algorithm were created, such as A* and Dijkstra's algorithm with Fibonacci heap \cite{hart1968formal}.





Motion planning is a specific application of path planning regarding autonomous robots. Different types of approaches can be used in motion planning, like Grid-based search (which transform the environment in a grid-mesh) \cite{grid}, Interval-based search (similar to grid-based search it but uses space data instead of a grid) \cite{interval} and Reward-based (similar to a reinforcement learning in deep learning) \cite{reward}. Commonly, path planning using Grid-based and Interval-based presents some issues in a large or dynamic environment, due to the time complexity of these algorithms and the need to recalculate the route if a new object was found in an already mapped space \cite{chen2016motion}.  





An important reason for the time complexity problem in the motion planning is because there are usually a lot of possible ways to get to the destination. Regarding the topic of information reducing, some algorithms of decomposition are applied in big data problems: Principal component analysis (PCA) \cite{pca} and Truncated Singular value decomposition (TSVD) \cite{tsvd}, Non-negative matrix factorization (NMF) \cite{nmf}, among others. Still, recently, these same solutions are used to reduce the number of possible ways to find the best route between two points \cite{pcapp2,pcapp1,svdpp}. However, many of the solutions proposed to compact data are not able to do that considering the non linearity of the data. In 2006 \cite{Hinton504}, Hinton, G.  proposed the use of Autoencoders to reduce the dimensionality of data, he showed how it is possible to represent data with much less information, because his proposal considered the non-linearity data. Since then, Autoencoders have been used to compress data, helping to solve various machine learning problems \cite{ae2}. 


In this paper, we investigate the application of a Convolutional Neuronal Network Encoder to reduce the number of routes. Therefore, the combination between our proposal and any algorithm of path planning must spend less iterations to find the best path than conventional A*. Moreover, we also propose a new dataset for evaluating the performance of the proposed method.

\section{CNN Encoder}
\label{secao_2}

Convolutional Neural Networks have an excellent capability to extract high-level features. For this reason, currently, it is being used to solve many problems in deep learning and computational vision problems\cite{lecun2015deep}. Simultaneously, Autoenconders are being used to code data information in unsupervised learning \cite{firstae}. They are trained to reconstruct the input data using fewer data than the original input; this way, many times, they can eliminate useless information. 


Thinking about the good results of CNNs and the capability of the Autoenconder to reduce the dimensionality of the data, we propose a CNN Encoder to eliminate useless routes from 2D maps. 

\subsection{Model architecture}
The process of building the architecture was made interactively. After many combinations of parameters and applications of transfer learning of famous CNNs, like VGG16 and  Resnet50, which gets a fewer number of iterations, can be seen in table \ref{arq}.
\begin{table}[h!]
\centering
\caption{Architecture of the CNN Encoder}
\label{arq}
\scalebox{0.8}{%
\begin{tabular}{|c|c|c|c|c|c|c|}
\hline
\multicolumn{2}{|c|}{Layer} & Filters & Kernel Size & Activation & Batch Norm & Dropout \\ \hline
1         & Image           & -       & -           & -          & -          & -       \\ \hline
2         & Conv            & 64      & 3x3         & ReLu       & True       & -       \\ \hline
3         & Conv            & 128     & 3x3         & ReLu       & False      & 30\%    \\ \hline
4         & Max-Pool        & -       & 3x3         & -          & -          & -       \\ \hline
5         & Conv            & 256     & 3x3         & ReLu       & True       & -       \\ \hline
6         & Conv            & 512     & 3x3         & ReLu       & False      & 30\%    \\ \hline
7         & Dense           & 256     & -           & LeakyReLU  & True       & 30\%    \\ \hline
8         & Dense           & 512     & -           & LeakyReLU  & True       & 30\%    \\ \hline
9         & Dense           & 1024    & -           & LeakyReLU  & True       & 30\%    \\ \hline
10        & Dense           & 3600    & -           & Tanh       & False      & 30\%    \\ \hline
\end{tabular}}
\end{table}

\section{Motion Planning Database}
\label{secao_3}
\begin{figure*}[h!]
\centering
 \subfloat[Example scene 1]{
   \includegraphics[width=.2\textwidth]{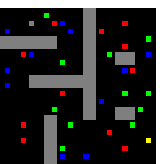}
   \label{a}
 }
 \qquad
 \subfloat[Example scene 2]{
   \includegraphics[width=.2\textwidth]{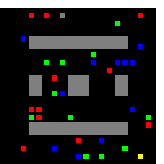}
   \label{b}
 }
  \qquad
 \subfloat[Example scene 3]{
   \includegraphics[width=.2\textwidth]{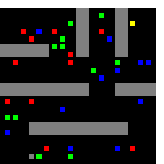}
   \label{c}
 }
  \qquad
 \subfloat[Example scene 4]{
   \includegraphics[width=.2\textwidth]{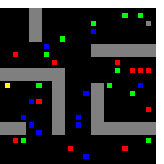}
   \label{d}
 }
  \qquad
 \subfloat[Example scene 5]{
   \includegraphics[width=.2\textwidth]{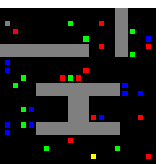}
   \label{e}
 }
 \caption{The instances of proposed the database. The yellow and gray pixels, respectively, represents the start and the end of the path.}
\label{database}
\end{figure*}

To check the efficiency of our solution and compare it to the conventional technique, we created an image database containing different scenarios, obstacles and goals. Figure \ref{database} shows instances of all proposed scenes. With our database, it is possible to evaluate the generalization capability of our model, applying our proposed approach in different situations and scenarios.
Our database contains five different scenarios, each of them composed of 10000 scenes (RGB images with $60\times60$ pixels), where each one has some fixed obstacles and others that are arranged randomly over the rest of the scene. Also, for each of these, there is an image answer, representing the expected path, created based on grid search technique.   

\section{Experimental results}
\label{secao_4}

\begin{figure*}[!h]
\centering
\includegraphics[width=.7\textwidth]{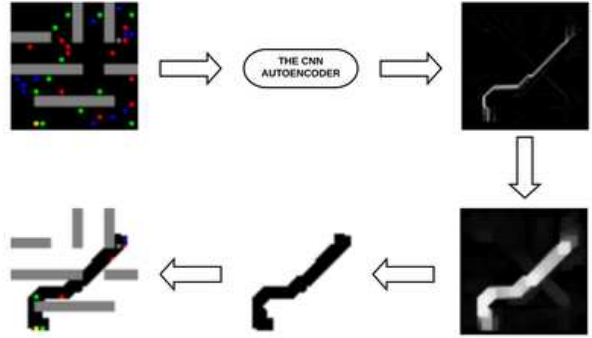}
\caption{The output processing of the CNN Encoder and comparison with the original input image. All black pixels in the first and last images represent the walkable paths.}
\label{diagrama}
\end{figure*}

To evaluate our work, we compare the number of iterations of A*, and the CNN Encoder (our proposal) with A*. The best solution is the one which found the best route with the least number of iterations.
The metrics that we used to evaluate the obtained results, how we split the database, the architecture, and the output processing are as following:
\begin{itemize}
    \item Metrics: Number of iterations. Represents the number of attempts until the algorithm found the shortest path.
    \item Split database: The database was split by percent split validation, 80\% to train set, and 20\% to test set.
    \item Output processing: The output processing has 5 steps, all of them can be seen in the Figure \ref{diagrama}. They are:
    \begin{enumerate}
        \item To Transform the output of the CNN Encoder (3600 values) to a Gray scale image with size 60x60.
        \item To Apply dilatation technique with kernel size equal to 3x3.
        \item To Binarize the image with threshold equal to 50.
        \item To Overlap the original scene and the processed output.
        \item To Apply the A* to find the shortest path, and to count the number of iterations.
    \end{enumerate}
    
\end{itemize}

\begin{table}[h!]
\centering
\caption{Comparison between the number of iterations of A* and the number of iterations of CNN Encoder + A*}
\label{results}
\scalebox{0.8}{%
\begin{tabular}{|l|l|l|l|l|l|l|}
\hline
            & Scene 1 & Scene 2 & Scene 3 & Scene 4 & Scene 5 & Sum      \\
\hline            
A*          & 3401942 & 2756298 & 1797817 & 2382324 & 2478137 & 12816518 \\
\hline
Proposal+A* & 1006854 & 911499  & 855623  & 890303  & 915310  & 4579589  \\
\hline
Difference & 2395088 & 1844799 & 942194  & 1492021 & 1562827 & 8236929  \\
\hline
Difference(\%)  & 29,60\% & 33,07\% & 47,59\% & 37,37\% & 36,94\% & 35,73\%  \\
\hline
Improvement & 70,40\% & 66,93\% & 52,41\% & 62,63\% & 63,06\% & 64,27\% \\
\hline
\end{tabular}}
\end{table}

Analyzing the results of Table \ref{results}, we can see that in all database scenarios, the number of iterations of the proposed solution presents an improvement of over 60\% compared to the conventional A*, which means that the number of iterations decreased considerably.

\begin{figure*}[h!]
\centering
 \subfloat[The paths taken by A* until found the shortest path]{
   \includegraphics[width=.8\textwidth]{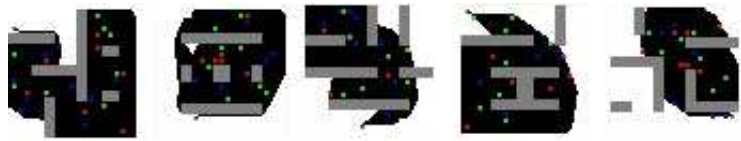}
   \label{a}
 }
 \qquad
 \subfloat[The paths taken by The CNN Enconder and A* until found the shortest path]{
   \includegraphics[width=.8\textwidth]{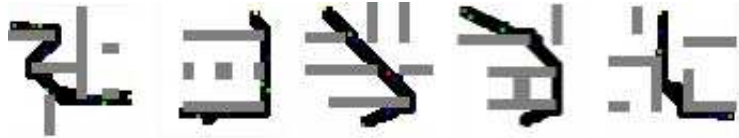}
   \label{d}
 }
 \caption{These images illustrate output of each explored algorithm.}
 \label{instances}
\label{results_img}
\end{figure*}

Observing the Image \ref{results_img} note which the A* many times try to find the shortest path using paths completely wrong, including going to the opposite side of the goal, it makes which the solution spend much time. On another hand, after the application of our proposal, the A* find the shortest path using fewer iterations. The CNN Encoder went almost directly to the goal and found that in much fewer iterations. 



\section{Conclusion}
\label{secao_5}

This work aimed to show the capabilities of a new algorithm based on the deep learning encoder to eliminate useless paths for motion planning algorithms. 

From the obtained results, we can assume that the proposed CNN architecture was able to learn to avoid of fixed and dynamic obstacles regardless of the presented scenario. This way, it is possible for other researchers to use our solution and our methods to improve their results on autonomous navigation issues. As such, our contribution is the CNN Enconder architecture, the database, and database creation methods using random obstacles, which are useful for Encoder learning process to motion planning tasks.

As future works, we hope to compare the efficiency of our solution with other motion planning algorithms, and data reduction techniques combined with path planning. Furthermore, we intend to study the computational cost of the proposed solution and other solutions in the literature, because many robotic solutions have limited hardware.


\begin{footnotesize}

\bibliographystyle{unsrt}
\bibliography{references}

\end{footnotesize}


\end{document}